# Multiscale Adaptive Representation of Signals: I. The Basic Framework

**Cheng Tai**                          CHENGT@MATH.PRINCETON.EDU
*PACM, Princeton University*
*Princeton, NJ 08544, USA*

**Weinan E**                             WEINAN@MATH.PRINCETON.EDU
*School of Mathematical Sciences and BICMR*
*Peking University and*
*Department of Mathematics and PACM*
*Princeton University*
*Princeton, NJ 08544, USA*

**Editor:**

## Abstract

We introduce a framework for designing multi-scale, adaptive, shift-invariant frames and bi-frames for representing signals. The new framework, called AdaFrame, improves over dictionary learning-based techniques in terms of computational efficiency at inference time. It improves classical multi-scale basis such as wavelet frames in terms of coding efficiency. It provides an attractive alternative to dictionary learning-based techniques for low level signal processing tasks, such as compression and denoising, as well as high level tasks, such as feature extraction for object recognition. Connections with deep convolutional networks are also discussed. In particular, the proposed framework reveals a drawback in the commonly used approach for visualizing the activations of the intermediate layers in convolutional networks, and suggests a natural alternative.

## 1. Introduction

It is now well acknowledged that sparse and overcomplete representations of data play a key role in many signal processing applications. The ability to represent a signal as a sparse linear combination of a few atoms from a possibly overcomplete dictionary lies at the heart of many applications including image/audio compression, denoising, as well as higher level tasks such as object recognition.

One popular technique for representing signals is the use of dictionaries. Since the seminal work of Olshausen and Field Olshausen et al. (1996), the field of dictionary learning has seen many promising advances. The objective is to learn a dictionary such that the input data can be written as a sparse linear combination of the dictionary atoms. More specifically, given the data represented as a matrix $X$, one finds the dictionary matrix $D$ and coefficient matrix $C$ simultaneously by solving:

$$\min_{D,C} \|X - DC\|_2^2 + \lambda \|C\|_1. \tag{1}$$

The solution is usually obtained by solving alternatively the minimization problem for $D$ and the sparse coding problem for $C$ with the other variable being kept fixed. After obtaining $D$, inference can be made by solving a sparse coding problem. Different dictionary learning models differ in the way the dictionary $D$ is updated. Examples include: MOD Engan et al. (1999a), K-SVD Aharon et al. (2006) and their variants.

Dictionary learning techniques have been successfully applied to some low level image and video processing tasks, such as image/video denoising Elad and Aharon (2006), compression Bryt and Elad (2008a); Engan et al. (1999b), inpainting Mairal et al. (2008) and other restoration tasks Mairal et al. (2007), with the state-of-the-art performances. In addition, dictionary learning and sparse coding techniques have been very popular in high level object recognition tasks where their function is to extract features from raw data. These techniques have been used successfully to extract visual features in Ranzato et al. (2007); Lee et al. (2009); Jarrett et al. (2009).

At the other end are the more traditional methodologies of designing analytic tight frames, such as Fourier basis, wavelet frames and bi-frames Daubechies et al. (2003), curvelets Candes and Donoho (2000), contourlets Do and Vetterli (2002), etc. These analytic tight frames are robust, easy to use and computationally efficient.

In some sense the analytic tight frames can also be viewed as a dictionary. The set of signals is a particular space of functions. A dictionary is found that gives rise to the optimal representation and approximation of the signals in that function class. The resulted dictionary is highly structured, and in particular, when used into applications, the dictionary atoms are never explicitly used. However, the two approaches do differ fundamentally in several aspects (see Table 1).

- **Computational cost.** For dictionary learning, the computational cost consists of two parts: the one time cost of learning the dictionary atoms and the repeated cost of solving the sparse coding problem for the test signal at inference time. Among the two, it is the latter that prevents it from being used in real time situations. Despite the efforts devoted to seeking more efficient sparse coding algorithms, see e.g. Daubechies et al. (2004); Lee et al. (2006); Beck and Teboulle (2009), none of the available techniques is efficient enough for large scale visual feature extraction. In fact, assuming that the signal $x$ is of length $N$ and the trained dictionary $D \in \mathbb{R}^{m \times N}$ is stored and used explicitly, then computing $Dx$ alone requires $O(mN)$ operations. In comparison, analytic transforms are far more efficient: fast Fourier transform takes $O(N \log N)$ operations and one level wavelet transform takes only $O(N)$ operations. This is a huge efficiency gap. In addition, the computational cost of training cannot be ignored either. The learning procedure requires solving a non-convex optimization problem, limiting dictionary atoms to low dimensions. Partly because of this, in image processing applications, dictionary atoms are only obtained for small image patches.

- **Multi-scale features.** Dictionaries as obtained by MOD and K-SVD operate at a single small scale. Since the dictionary atoms are limited to small sizes, there is not much room for multi-scale features. Past experience with wavelets has taught us that often times it is beneficial to process signals at several scales, and operate at each scale separately.

- **Artifacts.** In low level tasks such as image compression, the dictionary learning approach operates in a patch by patch manner, which produces visually unpleasant block effects along the boarders of the patches Bryt and Elad (2008a). Post processing is often needed to remove these artifacts Bryt and Elad (2008b).

|  | Dictionary Learning | Wavelet Tight Frames |
|---|---|---|
| Adapted to data | **Yes** | No |
| Computational speed | Slow | **Fast** |
| Multi-scale | No | **Yes** |
| Robustness to perturbation | Conditionally | **Yes** |
| Performance on real data | **Better** | Worse |

Table 1: Comparison between dictionary learning and wavelet tight frames

Given the relative features of dictionary learning and wavelet tight frames, it is natural to ask whether one can design bases that have the benefits of both and avoid the problems. In other words, can one design bases that are adapted to the data but at the same time have the multi-scale structure that is essential for the efficient algorithms for wavelet tight frames?

We propose a framework of constructing adaptive frames and bi-frames (abbreviated as AdaFrame). This framework gives multi-scale, sparse representations of the signal, with an efficiency comparable to that of the wavelets at inference time.

The proposed framework is formally similar to the first few layers of a convolutional network. As a byproduct, we show that the proposed framework gives a better way of visualizing the activations of the intermediate layers of a neural net in terms of reconstruction error.

The framework presented here is best suited for datasets such that each data point has some structure. Obvious examples include time series, images and videos. However, as in the case of wavelets, it is also possible to extend this kind of ideas to less structured data such as graphs, etc Coifman and Maggioni (2006).

Most examples discussed in this paper are still of the low level image processing type. In a subsequent paper, we will discuss more thoroughly higher level tasks such as image classification.

The organization of this paper is as follows. In section 2, we introduce shift-invariant frames and bi-frames. In section 3, we introduce the adaptive construction of shift-invariant frames. In section 4, we introduce the adaptive construction of shift-invariant bi-frames. In section 5, we discuss multi-level constructions. In section 6, we give some simple illustrative examples of the adaptively constructed frames and bi-frames. In section 7, we discuss the connection with predefined wavelets and wavelet frames. In section 8, we discuss applications to image processing and image classification. In section 9, we discuss connection with deconvolutional nets and reconstruction of input data from features in the intermediate layers of the convolutional nets. Some conclusions are drawn in section 10.

## 2. Shift-invariant Frames and Bi-frames

An important starting point is the concept of multi-resolution analysis (MRA) introduced by Mallat Mallat (1989) and Meyer Meyer (1995), of which wavelets are particularly popular examples. One main advantage of MRA is that it comes naturally with fast decomposition and reconstruction algorithms, and this has been essential for making wavelets a practical tool in signal processing Daubechies et al. (2003); Shen (2010). Although our work builds upon the theory of wavelet frames in the continuous setting, we decide to introduce our model in a purely discrete setup. This has the advantage that it is more direct and more easily linked with existing machine learning models, including dictionary learning and convolutional networks. However, as noted in Han (2010), there is a canonical link between affine systems in the continuous setting and fast algorithms in the discrete framework.

The signals and the filters are all assumed to be discrete sequences in $l_2(\mathbb{Z}^d)$, where $d$ is the dimension. For audio, image and video signals, $d = 1, 2, 3$ respectively. First let us define the up- and down-sampling operators. Let $M$ be an integer. The (one dimensional) down-sampling and up-sampling operator are defined by:

$$[v \downarrow_M](n) := v(Mn), \quad n \in \mathbb{Z}$$
$$[\uparrow v](n) := \begin{cases} v(k), & n = Mk, k \in \mathbb{Z} \\ 0, & \text{otherwise} \end{cases} \quad (2)$$

respectively, for $v \in l_2(\mathbb{Z})$. $M$ is the decimation factor. Similarly if $d > 1$, denote the decimation factor in each dimension by $M_1, M_2, \cdots, M_d$. For convenience we define a matrix $M = \text{Diag}(M_1, \cdots, M_d) \in \mathbb{R}^{d \times d}$. A common choice of $M$ in image processing is $M = 2I$. We call $M$ the *sampling matrix* and use the same notation as in (2) where $Mn$ is understood as the matrix-vector multiplication. In general $M$ can be an invertible matrix whose entries are positive integers or rational numbers that are greater than 1.

Key to the decomposition and reconstruction algorithms are the transition and subdivision operators. For a data sequence $v \in l_2(\mathbb{Z}^d)$, a finitely supported filter $a \in l_2(\mathbb{Z}^d)$ and a sampling matrix $M \in \mathbb{R}^{d \times d}$, the transition operator $\mathcal{T}_a : l_2(\mathbb{Z}^d) \mapsto l_2(\mathbb{Z}^d)$ is defined by

$$(\mathcal{T}_{a,M} v)(n) := \downarrow_M [v * a](n) = \sum_{k \in \mathbb{Z}^d} v(k) \overline{a(k - Mn)}, \qquad (3)$$

the subdivision operator $\mathcal{S}_a : l_2(\mathbb{Z}^d) \mapsto l_2(\mathbb{Z}^d)$ is defined by

$$(\mathcal{S}_{a,M} v)(n) := |\det(M)| [a * (\uparrow v)](n) = |\det(M)| \sum_{k \in \mathbb{Z}^d} v(k) a(n - Mk). \qquad (4)$$

To make the notations more concise, we omit $M$ in the subscript.

Given a set of finitely supported filters $A = \{a_1, \cdots, a_m\}$ and the coefficient sequence $v \in l_2(\mathbb{Z}^d)$, which could be the input signal itself or the coefficients computed at some decomposition level, we compute coefficients of the next level by

$$v_l = \mathcal{T}_{a_l} v, \quad l = 1, \cdots, m. \qquad (5)$$

With this notation, the one-level decomposition operator $W_A : l_2(\mathbb{Z}^d) \mapsto \underbrace{l_2(\mathbb{Z}^d) \oplus \cdots \oplus l_2(\mathbb{Z}^d)}_{m \text{ times}}$

is defined as:
$$W_A v := \{v_1, \cdots, v_l\} = \{\mathcal{T}_{a_1} v, \mathcal{T}_{a_2} v, \cdots, \mathcal{T}_{a_m} v\}. \qquad (6)$$

Given a set of finitely supported filters $B = \{b_1, \cdots, b_m\}$, the one-level reconstruction operator $R_b : \underbrace{l_2(\mathbb{Z}^d) \oplus \cdots \oplus l_2(\mathbb{Z}^d)}_{m \text{ times}} \mapsto l_2(\mathbb{Z}^d)$ is defined as

$$R_B(v_1, \cdots, v_m) := \sum_{l=1}^{m} \mathcal{S}_{b_l} v_l \qquad (7)$$

In wavelet frames, the filters $A$ used for decomposition and the filters $B$ used for reconstruction are connected by : $b_l(\cdot) = a_l(-\cdot), l = 1, \cdots, m$, where $a_l(-\cdot)$ means flip the entries of $a_l$ along each dimension. But this does not have to be the case: $A$ and $B$ can be different and together they constitute a *bi-frame*.

The main requirement is that of perfect reconstruction, by which we mean:

$$R_B W_A v = v \quad \forall v \in l_2(\mathbb{Z}^d). \qquad (8)$$

The following result is crucial.

**Theorem 1** *Daubechies et al. (2003) Let $M \in \mathbb{R}^{d \times d}$ be a sampling matrix, let $A = \{a_1, \cdots, a_m\}$ and $B = \{b_1, \cdots, b_m\}$ be two sets of finitely supported sequences in $l_2(\mathbb{Z}^d)$. Then the perfect reconstruction property*

$$R_B W_A v = v, \quad \forall v \in l_2(\mathbb{Z}^d) \qquad (9)$$

holds if and only if, for all $k, j \in \mathbb{Z}^d$,

$$\sum_{l=1}^{m} \sum_{n \in \mathbb{Z}^d} a_l(Mn+j)\overline{b_l(k+Mn+j)} = |\det(M)|^{-1}\delta_k \tag{10}$$

where $\delta_k = 1$ if $k = 0$ and $\delta_k = 0$ otherwise.

In the case of wavelet tight frames, $b_l(\cdot) = a_l(-\cdot), l = 1, \cdots, m$, and we have:

**Theorem 2** *Daubechies et al. (2003) Let $M \in \mathbb{R}^{d \times d}$ be a sampling matrix, let $A = \{a_1, \cdots, a_m\}$ be a set of finitely supported sequences in $l_2(\mathbb{Z}^d)$. Then the perfect reconstruction property*

$$R_A W_A v = v, \quad \forall v \in l_2(\mathbb{Z}^d). \tag{11}$$

*holds if and only if, for all $k, j \in \mathbb{Z}^d$,*

$$\sum_{l=1}^{m} \sum_{n \in \mathbb{Z}^d} a_l(Mn+j)\overline{a_l(k+Mn+j)} = |\det(M)|^{-1}\delta_k \tag{12}$$

In particular, if the data are real numbers and no down-sampling is performed, then the perfect reconstruction condition (12) becomes

$$\sum_{i=1}^{m} \sum_{n \in \mathbb{Z}^d} a_i(k+n)a_i(n) = \delta_k, \forall k \in \mathbb{Z}^d. \tag{13}$$

The proof of Theorem 1 and Theorem 2 can be found in Daubechies et al. (2003). For completeness, we give a direct proof for the discrete case in the appendix. These conditions are referred to as the unitary extension principle (UEP) in wavelet frame theory.

As an example, the linear B-spline wavelet tight frame used in many image restoration tasks is constructed via the UEP. Its associated filters are :

$$a_1 = \frac{1}{4}(1, 2, 1)^T; \quad a_2 = \frac{\sqrt{2}}{4}(1, 0, -1)^T; \quad a_3 = \frac{1}{4}(-1, 2, -1)^T.$$

This kind of tight frames are shift-invariant systems since the transforms are in the form of discrete convolution. They are suited for the case when, below certain scale, the statistical properties of the signals are translation invariant.

## 3. Adaptive Construction of Frames

Given a set of signals $X = \{x_1, \cdots, x_N\}$, the goal is to construct wavelet frames that are adapted to this set of signals in the sense that signals in the given set have a sparse representation.

Define $\mathcal{Q}$ to be the set of filters that satisfy the UEP condition:

$$\mathcal{Q} = \left\{ \{a_i\}_{i=1}^m : \sum_{l=1}^m \sum_{n \in \mathbb{Z}^d} a_l(Mn+j)\overline{a_l(Mn+k+j)} = |\det(M)|^{-1}\delta_k, \ \forall k, j \in \mathbb{Z}^d \right\}. \tag{14}$$

Filters in this set generate a wavelet frame that provide a faithful representation for all signals in $l_2(\mathbb{Z}^d)$. However, we are not interested in all signals in $l_2(\mathbb{Z}^d)$. We are only interested in $X$. Among all filters in $\mathcal{Q}$, we want to select the one that is most adapted to $X$.

In image restoration tasks, we are mostly interested in wavelet frames that give rise to a sparse representation of the input signal. Therefore we will use sparsity as our guiding principle for selecting the filters. Other guiding principles such as the discriminative criterion can also be used. But in this paper, we will focus on sparsity.

Let $\Phi$ be a sparsity-inducing function. Examples of $\Phi$ include the $l_1$ norm, $l_0$ "norm", or the Huber loss function defined (component-wise) by:

$$L_\delta(x) = \begin{cases} \frac{1}{2}x^2, & |x| \leq \delta \\ \delta(|x| - \frac{1}{2}\delta), & \text{otherwise} \end{cases}. \tag{15}$$

Given the data $X$, the adaptive filters are chosen by solving the following optimization problem:

$$\begin{aligned} \min_{a_1,\cdots,a_m} & \sum_{j=1}^N \sum_{i=1}^m \Phi(v_{i,j}) \\ \text{subject to} \quad & v_{i,j} = \mathcal{T}_{a_i} x_j, \quad i = 1, \cdots, m \\ & \{a_i\}_{i=1}^m \in \mathcal{Q} \end{aligned} \tag{16}$$

In the following, without loss of generality, we will assume that there is only one data point in the signal set, i.e. $N = 1$, and we will omit the subscript $j$.

To be specific, we use $l_1$ norm as the measurement of sparsity and we will note the changes required if the $l_0$ norm is used. The above problem then becomes

$$\begin{aligned} \min_{a_1,\cdots,a_m} & \sum_{i=1}^m \|\mathcal{T}_{a_i} x\|_1 \\ & \{a_i\}_{i=1}^m \in \mathcal{Q} \end{aligned} \tag{17}$$

This innocent looking optimization problem is difficult to solve because of the constraint. Consider the simplest case when the signals and the filters are all one-dimensional. Assume each filter has support length $r$, and we have $r$ of them. For a real symmetric matrix $G$, let us denote by $Tr(G, k)$ the sum of entries along the $k$-th sub-diagonal. For example, $Tr(G, 0)$ is the usual trace of $G$. Let $A := (a_1, \cdots, a_m)$. Then the constraint $\{a_i\}_{i=1}^m \in \mathcal{Q}$ is equivalent to

$$Tr(AA^T, k) = \delta_k, \ k = 0, \cdots, r-1.$$

To see a nontrivial example where this constraint is satisfied, take an orthorgonal matrix $U \in \mathbb{R}^{r \times r}$, and let $a_i = \frac{1}{\sqrt{r}} U_{\cdot,i}, i = 1, \cdots, m$, where $U_{\cdot,i}$ means the $i$-th column of $U$. However, in general, the algebraic constraint above is difficult to deal with. Note also that this optimization problem is not convex.

We use the split Bregman algorithm Goldstein and Osher (2009) to solve (17). Introduce the auxiliary variable $D = (d_1, \cdots, d_m)$ where $d_i = \mathcal{T}_{a_i} x, i = 1, \cdots, m$. Define the norm $\|D\|_{1,1} := \sum_{i=1}^m \|d_i\|_1$. Then (17) is equivalent to:

$$\begin{aligned} \min_{A, D} \quad & \|D\|_{1,1} \\ \text{subject to} \quad & D = W_A x \\ & A \in \mathcal{Q} \end{aligned} \quad (18)$$

Applying the split Bregman method, we obtain the following algorithm:

---
**Algorithm 1** Adaptive construction of frames
---
1: **Input:** $x$.
2: **Initialize** $k = 0, B = \mathbf{0}, A = A^0, D = W_{A^0} x$.
3: **while** "not converge" **do**
4: $\quad D^{k+1} \leftarrow \arg\min_D \|D\|_{1,1} + \frac{\eta}{2} \|D - W_{A^k} x - B^k\|_F^2$
5: $\quad A^{k+1} \leftarrow \arg\min_A \|W_A x - D^{k+1} + B^k\|_F^2$ **s.t.** $A \in \mathcal{Q}$.
6: $\quad B^{k+1} \leftarrow B^k + W_{A^{k+1}} - D^{k+1}$.
7: $\quad k \leftarrow k + 1$
8: **return** $A^k$
---

To implement the algorithm, we must be able to solve each of the subproblems listed in steps 4, 5 and 6.

To solve the subproblem for $D$, note that the problem decouples for each $d_i, i = 1, \cdots, m$. In fact,

$$d_i^{k+1} = \arg\min_d \left( \|d\|_1 + \frac{\eta}{2} \|\mathcal{T}_{a_i^k} x - d + b_i^k\| \right) \quad (19)$$

for $i = 1, \cdots, m$. It is easy to see that (19) has a closed form solution given by

$$d_i^{k+1} = \text{shrink}(\mathcal{T}_{a_i^k} x + b_i^k, \frac{1}{\eta}) \quad (20)$$

where the function shrink : $\mathbb{R} \mapsto \mathbb{R}$ is defined as

$$\text{shrink}(x, a) = \begin{cases} (|x| - a)sign(x), & \text{if } |x| > a \\ 0, & \text{otherwise} \end{cases}. \tag{21}$$

When shrinkage-operator acts on a vector, it acts on each component of the vector according to (21).

The subproblem for updating $A$ is most problematic due to the constraint. We use the interior-point method for this part of the algorithm. There is no guarantee of a global solution to this subproblem.

The update for $B$ is straightforward. This is analogous to the step of "adding the noise back" in the ROF model for denoising Osher et al. (2005).

Among the three subproblems, the update of $A$ is the most time consuming. But as is observed by many authors, it is not necessary to solve $A$ to full convergence, the intuitive reason being that if the error of the solution to the subproblem is smaller than $\|B^k - B^{k-1}\|$, the extra accuracy will be wasted. In fact, for updating $A$, we only run a few steps of the interior-point iterations and we still observe numerical convergence.

If we use the $l_0$ "norm" as the measurement of sparsity, the only change needed in the above algorithm is in the $D$ step, where the soft-shrinkage operator is replaced by hard-thresholding defined as:

$$\text{Hard}(x, a) = \begin{cases} x, & \text{if } |x| > a \\ 0, & \text{otherwise} \end{cases}. \tag{22}$$

To give the readers some intuition about how the filters obtained look like, we show an example in Figure 1. More examples are given in section 5.

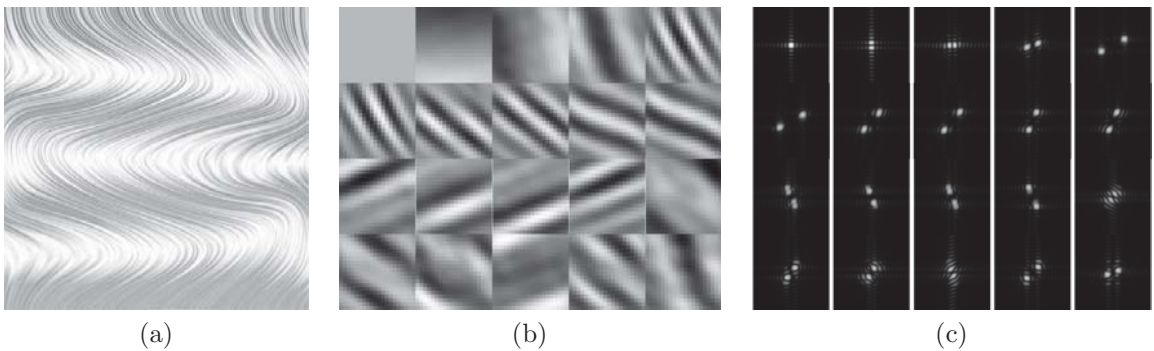

(a) (b) (c)

Figure 1: (a) The input image.(b) The filters learned using Algorithm 1, $m = 20, r = 20$.(c) The Fourier spectrum of the corresponding filters. Note the first filter is a low-pass filter, all other filters are high-pass filters as can be seen from the Fourier spectrum. The second and third filter look like edge detectors along the axis. Other filters detect oscillations along different directions.

A special case of this model in 2D, where the support of $a_i$ is of size $r \times r$, $m = r^2$ and the filters are orthogonal to each other was proposed in Cai et al. (2014). A local solution was found there by alternating between thresholding and singular value decomposition.

In some applications such as object recognition, perfect reconstruction is unnecessary. Instead, writing the input signal as a sparse linear combination of a few dictionary atoms is only a means to extract features to be used by other learning algorithms. Sparse coding has been quite popular in serving this purpose for visual object recognition tasks. In this case, it is possible to relax the constraint in (17). Instead of solving the constrained minimization problem, we can use a penalty method to solve an unconstrained problem. For example, in 1D, we can solve

$$\min_A \sum_{i=1}^m \|W_A x\|_{1,1} + \eta \sum_k (Tr(AA^T, k) - \delta_k)^2 \qquad (23)$$

where $\eta$ is a parameter that depends on our tolerance on the reconstruction error. This unconstrained problem is relatively easy to solve using first-order optimization methods.

## 4. Adaptive Construction of Bi-frames

In this section, we introduce the adaptive construction of wavelet bi-frames. Compared with the wavelet frames, the bi-frames offer two distinct advantages: The first is that the constraint for the filters becomes bi-linear making it easier to construct the filters. The second is that the added redundancy introduces more flexibility. These prove to be very important in practice.

Let $\mathcal{Q}$ denote the set of pairs $A$ and $B$, $A = (a_1, \cdots, a_m), B = (b_1, \cdots, b_m)$, that satisfy (10):

$$\mathcal{Q} := \left\{ (A, B) : \sum_{l=1}^m \sum_{n \in \mathbb{Z}^d} a_l(Mn+j)\overline{b_l(k+Mn+j)} = |\det(M)|^{-1}\delta_k,\ \forall k, j \in \mathbb{Z}^d \right\} \qquad (24)$$

We want to find filter pairs $(A, B)$ with desired properties while respect the constraint $(A, B) \in \mathcal{Q}$. As before we will only consider sparsity. Given the data $x$ and a sampling matrix $M$, we aim to solve :

$$\begin{aligned} \min_{A,B} \quad & \|W_A x\|_{1,1} \\ \text{subject to} \quad & (A, B) \in \mathcal{Q} \end{aligned} \qquad (25)$$

The constraint $(A, B) \in \mathcal{Q}$ in bi-linear in $A$ and $B$. Let us first count the number of equations.

We start with the simplest case where the signals and the filters are one dimensional. Let $A, B$ be defined as before and assume that each filter $a_i, b_i, i = 1, \cdots, m$ has support size $r$. Given the decimation factor $M$, define

$$\mathcal{S}(r) := \{(k, \gamma) : \exists n \in \mathbb{Z}, 1 \leq Mn + k + \gamma \leq r, 1 \leq Mn + \gamma \leq r\}, \quad (26)$$

then each $(k, \gamma) \in \mathcal{S}(r)$ constitutes an equation. This gives

$$|S(r)| = (2r - M)M. \quad (27)$$

This is the total number of equations. The total number of unknowns in $A$ and $B$ is $2rm$. Therefore for (10) to have a solution, we expect:

$$2rm \geq (2r - M)M. \quad (28)$$

In the general case where the signals and the filters live in $d$ dimensions, we can do a similar counting. Assume the support size of the filter $a_i, b_i, i = 1, \cdots, m$ is $\mathbf{r} = (r_1, \cdots, r_d)$, and assume that the sampling matrix is $M = \text{Diag}(M_1, \cdots, M_d)$. Let

$$\mathcal{S}(r) := \{(k, \gamma) \in \mathbb{Z}^d : \exists n \in \mathbb{Z}^d, \mathbf{1} \leq Mn + k + \gamma \leq \mathbf{r}, \mathbf{1} \leq Mn + \gamma \leq \mathbf{r}\}, \quad (29)$$

where the inequality is understood component-wise. Each $(k, \gamma) \in \mathcal{S}(r)$ gives rise to an equation. The total number of equations is

$$|S(r)| = \prod_{i=1}^{d} (2r_i - M_i) M_i. \quad (30)$$

The number of unknowns in $a$ and $b$ is $2m \prod_{i=1}^{d} r_i$. Hence to have a solution to (10), we expect:

$$2m \prod_{i=1}^{d} r_i \geq \prod_{i=1}^{d} (2r_i - M_i) M_i. \quad (31)$$

Two cases are of special interest.

- **Redundant case**. In this case, the number of filters $m$ is large. The number of decomposition coefficients is larger than the size of the input signal. Hence we call this the *redundant case*. For the optimization problem, we have more unknowns than equations. In particular, if $m \geq 2M - M^2/r$ in one dimension, and $m \prod_{i=1}^{d} r_i \geq \prod_{i=1}^{d} (2r_i - M_i) M_i$ in $d$ dimensions, for most $A$, we expect (10) as a set of linear equations for $B$, to have a solution. Therefore, we can design $A$ and $B$ separately: We can design $A$ first in whichever way we want as long as it is non-degenerate. We then solve (10) to get $B$.

- **Critically down-sampled case.** In this case, the number of filters $m$ is small. The number of decomposition coefficients is the same as that of the input signal (depending on the boundary conditions). Hence we call this the *critically down-sampled case*. For example, in one dimension, $m = M$. In this down-sampled case, for a typical $A$, it is likely that (10), as a linear system for $B$, does not have a solution. This means that we must consider $A$ and $B$ simultaneously.

## 4.1 Redundant Case

### 4.1.1 Design of the Decomposition Filters

As discussed above, we can design $A$ in the first phase, and then choose $B$ that satisfies the linear constraint (25) in the second phase.

However, the choice of $A$ has a significant impact on the condition number of (25). Hence some constraints should be added. While there are a lot of flexibilities, we propose the following formulation:

$$\begin{aligned} \min_{A} \quad & \|W_A x\|_{1,1} \\ \text{subject to} \quad & A^T A = I \end{aligned} \tag{32}$$

The additional constraint $A^T A = I$ is chosen based on the consideration that the filters are most incoherent among themselves.

To solve (32) numerically, we apply the split Bregman method. But we need to handle the extra orthogonality constraint as well. To this end, we introduce the auxiliary variable $P = A$ as a means to split the orthogonality constraint. This trick has been used in other problems, see for example Lai and Osher (2014). The problem then becomes:

$$\begin{aligned} \min_{A,D,P} \quad & \|D\|_{1,1} \\ \text{subject to} \quad & D = W_A x, P = A, P^T P = I. \end{aligned} \tag{33}$$

The algorithm is then:

---
**Algorithm 2** Adaptive construction of bi-frames: redundant case
---
1: **Input:** $x$.
2: Initialize $k = 0, F = \mathbf{0}, C = \mathbf{0}, A = A^0, D = W_{A^0} x, P = A$.
3: **while** "not converge" **do**
4:     **for** n=1:N **do**
5:         $D^{k+1} \leftarrow \arg\min_D \|D\|_{1,1} + \frac{\eta}{2}\|D - W_{A^k} x - F^k\|_F^2$
6:         $A^{k+1} \leftarrow \arg\min_A \eta\|W_A x - D^{k+1} + F^k\|_F^2 + \lambda\|A - P^k + C^k\|_F^2$
7:     $P^{k+1} \leftarrow \arg\min_P \|A^{k+1} - P + C^k\|_F^2 \quad$ **s.t.** $\quad P^T P = I$
8:     $F^{k+1} \leftarrow F^k + W_{A^{k+1}} x - D^{k+1}$.
9:     $C^{k+1} \leftarrow C^k + A^{k+1} - P^{k+1}$
10:    $k \leftarrow k + 1$
11: **return** $A^k$
---

To implement the algorithm, we must be able to solve each of the subproblems for $D, A$ and $P$. Updating $D$ is the same as in Algorithm 1. The subproblem for $A$ is a quadratic program. It can be decoupled into $m$ smaller problems, each of which involves one column of $A$. Writing $D = (d_1, \cdots, d_m)$, $P = (p_1, \cdots, p_m)$, $C = (c_1, \cdots, c_m)$ and $F = (f_1, \cdots, f_m)$, we can perform the optimization in a column by column fashion:

$$a_i^{k+1} = \arg\min_a \ \eta \|\mathcal{T}_a x - d_i^{k+1} + f_i^k\|_2^2 + \lambda \|a - p_i^k + c_i^k\|_2^2, \quad i = 1, \cdots, m \qquad (34)$$

Each of the $m$ smaller problems is an unconstrained quadratic program. Many optimization techniques can be used to to solve this problem. Among the several choices of iterative algorithms, we use conjugate gradient (CG) method because the objective function value tends to decrease very quickly in the first few CG iterations, thus giving a good approximate solution quickly. For the same reason as in Algorithm 1, iteration to convergence is not necessary.

Next, we consider the subproblem for $P$. This problem is equivalent to:

$$\max_P \ \text{Trace}((A^{k+1} + C^k)^T P) \quad \text{subject to} \quad P^T P = I. \qquad (35)$$

This is the classical orthogonal procrustes problem Gower and Dijksterhuis (2004) and has a closed form solution which we summarize in the following lemma. The proof can be found in the linear algebra textbooks, e.g. Horn and Johnson, chapter 3.

**Lemma 1** *Let $Y \in \mathbb{R}^{n \times m}, n \geq m$ and $Y = UDV^T$ be the singular value decomposition of $Y$, then the constrained optimization problem*

$$P^* = \arg\min_{P \in \mathbb{R}^{n \times m}} \|P - Y\|_F^2 \quad \text{subject to} \quad P^T P = I \qquad (36)$$

*has a closed form solution given by $P^* = U I_{n \times m} V^T$.*

Substituting $Y$ with $A^{k+1} + C^k$, we get the formula for updating $P$. Updating the auxiliary variable $F$ and $C$ is straightforward.

An illustration of such filters is shown in Figure 8(b).

4.1.2 Design of the Reconstruction Filters

Once $A = (a_1, \cdots, a_m)$ is obtained, we move on to second phase of designing the reconstruction filters $B$.

For fixed $A$ and sampling matrix $M$, the constraint (10) is a linear system in $B$. Hence we will write it as $H(A)B = f$, where $H(A)$ denotes the coefficient matrix generated using $A$. To get some concrete ideas, let us look at a simple example.

**Example.** Consider a one dimensional situation where $m = 2, r = 3$. Assume $A = (a_1, a_2), B = (b_1, b_2) \in \mathbb{R}^{3\times 2}$,

$$A = \begin{pmatrix} a_{11} & a_{21} \\ a_{12} & a_{22} \\ a_{13} & a_{23} \end{pmatrix}, \quad B = \begin{pmatrix} b_{11} & b_{21} \\ b_{12} & b_{22} \\ b_{13} & b_{23} \end{pmatrix},$$

Assume $M = 1$, that is, no downsampling is performed. Then the linear equation $H(A)B = f$ is

$$\begin{pmatrix} a_{11} & a_{12} & a_{13} & a_{21} & a_{22} & a_{23} \\ 0 & a_{11} & a_{12} & 0 & a_{21} & a_{23} \\ 0 & 0 & a_{11} & 0 & 0 & a_{21} \\ a_{12} & a_{13} & 0 & a_{22} & a_{23} & 0 \\ a_{13} & 0 & 0 & a_{23} & 0 & 0 \end{pmatrix} \begin{pmatrix} b_{11} \\ b_{12} \\ b_{13} \\ b_{21} \\ b_{22} \\ b_{23} \end{pmatrix} = \begin{pmatrix} 1 \\ 0 \\ 0 \\ 0 \\ 0 \end{pmatrix}.$$

This is a system of 5 equationd with 6 unknowns. Therefore, we have one additional degree of freedom left to design $B$.

In general, since $m$ is large, $H(A)B = f$ is an under-determined linear system. Moreover, since $A$ is obtained by solving (32) with respect to the orthogonality constraint, the coefficient matrix $H(A)$ tend to have a good condition number. This well-behaved under-determined linear system gives us the freedom to design the reconstruction filters $B$ with additional properties. The general formulation is:

$$\min_B \ G(B) \quad \text{subject to} \quad H(A)B = f \text{ and other constraints} \tag{37}$$

where $G(B)$ is the objective function that we use to impose the additional property that we expect $B$ to have. For example, if we want the reconstruction filters to look like piecewise smooth function, we can use the following formulation:

$$\begin{aligned} \min_B \quad & G(B) := \sum_{l=1}^m \|\nabla b_l\|_1 \\ \text{subject to} \quad & \|b_l\|_2 = \alpha, \ l = 1, \cdots, m \\ & H(A)B = f \end{aligned} \tag{38}$$

where $\nabla$ is a discrete gradient operator and $\alpha$ is a predefined parameter whose purpose is to make the size of $B$ compatible with the constraint $H(A)B = f$.

An illustration of the reconstruction filters is given in Figure 8(b).

### 4.2 Critically Down-sampled Case

In this case, we have less freedom and must consider the decomposition and reconstruction filters simultaneously. Since the constraint is bi-linear in $A$ and $B$, in order to avoid the trivial situation where the objective function is minimized by scaling down the decomposition filters $A$ and scaling up the reconstruction filters $B$, we require the filters $A$ to have unit norm. Adopting the same notation as before, (25) becomes

$$\begin{aligned}\min_{A,B} \quad & \|W_A x\|_{1,1} \\ \text{subject to} \quad & H(A)B = f. \\ & \|a_i\|_2 = 1, \ i = 1, \cdots, m\end{aligned} \qquad (39)$$

Again, we apply the split Bregman algorithm to solve this problem. The procedures are similar to the redundant case. We will formulate the algorithm directly as follows:

---
**Algorithm 3** Adaptive construction of bi-frames: critically down-sampled case

1: **Input:** $x$.
2: **Initialize** $k = 0, F = \mathbf{0}, C = \mathbf{0}, A = A^0, B = B^0, D = W_{A^0}x$.
3: **while** "not converge" **do**
4: $\quad D^{k+1} \leftarrow \arg\min_D \ \|D\|_{1,1} + \frac{\eta}{2}\|D - W_{A^k}x - F^k\|_F^2$
5: $\quad A^{k+1} \leftarrow \arg\min_A \ \eta\|W_A x - D^{k+1} + F^k\|_F^2 + \lambda\|H(A)B^k - f + C^k\|_F^2$ **s.t.** $\|a_i\|_2 = 1, i = 1, \cdots, m$
6: $\quad B^{k+1} \leftarrow \arg\min_B \lambda\|H(A^{k+1})B - f + C^k\|_F^2$
7: $\quad F^{k+1} \leftarrow F^k + W_{A^{k+1}}x - D^{k+1}$.
8: $\quad C^{k+1} \leftarrow C^k + H(A^{k+1})B^{k+1} - f - P^{k+1}$
9: $\quad k \leftarrow k+1$
10: **return** $A^k, B^k$

---

Updating $D$ is again done by soft thresholding. The update of $A$ is done by running a few iterations of the interior-point method, and updating $B$ is done by running a few iterations of conjugate gradient method. The most computationally intensive step is updating $A$. But since in our applications, the support size and the number of filters are small, the total number of variables is normally a few hundred, hence the computational cost is reasonable.

## 5. Multi-level Adaptive Frames

Going to multi-level, the basic idea is to recursively use the framework of adaptive frames on the coefficients obtained by applying the adaptive filters to the signal. There are two practical issues that we need to consider. The first is whether one considers all the coefficients or a subset of coefficients when going to coarser level. In this regard the difference between low-pass and high-pass filters is particularly relevant. Recall that a low pass filter is defined by the condition that the Fourier coefficient $\hat{a}(0) \neq 0$. The second issue is whether a new set of adaptive filters is learned and used at each level. We will discuss three different strategies that are motivated by three different examples.

## 5.1 The MRA approach

The basic idea of MRA is to apply the same set of filters at each level to the coefficients from the low-pass filters. When constructing traditional wavelet frames using MRA, there is only one low-pass filter at each level, the scaling function. All other filters are high pass filters associated with the wavelets. Our experience suggests that this is often the case for the adaptively learned filters. To makes sure that this is indeed the case, we can also add the additional constraint

$$\hat{a}_1(0) \neq 0, \hat{a}_i(0) = 0, i = 2, \cdots, m \qquad (40)$$

to (17). As a linear constraint, this does not cause much trouble in the optimization algorithm. With this, the adaptive wavelet frames can be used in the same way as classical wavelet frames. Specifically, given the the input signal $x$, the multi-level decomposition proceeds as follows: We first perform a one-level decomposition to get the coefficients $v_i = \mathcal{T}_{a_i} x, i = 1, \cdots, m$. $v_1$ is associated with the low-pass filter, which provides the coarse-grained approximation of the signal, and $v_i, i = 2, \cdots, m$ are associated with the high-pass filters, which provide the missing details from the coarse-graining. Next, we treat $v_1$ as the input signal and perform another one-level decomposition using the same set of filters to get the second-level coefficients. This procedure can then be continued. Schematically, this algorithm can be represented as a tree with one branching point at each level, as shown in Figure 2(a).

## 5.2 The scattering transform approach

By applying each fixed filter to the signal, one obtains a set of coefficients, called a feature map. If the input signal is an image, the feature map is also an image. One can then treat this new image as the input signal and find the corresponding adaptive filters. In some applications, this can be preceded by some component-wise nonlinear transformation. This is schematically shown in Figure 2(b). This structure is used in the scattering transforms proposed in Bruna and Mallat (2013).

The obvious drawback of this approach is that the degrees of freedom increase exponentially as the number of levels increases. Nevertheless, in classification tasks, it is generally believed that lifting the raw data to a high dimensional space using some nonlinear transforms can help by making the data more linearly separable. This is the underlying principle that makes kernel methods effective. Therefore this approach is potentially useful for classification tasks.

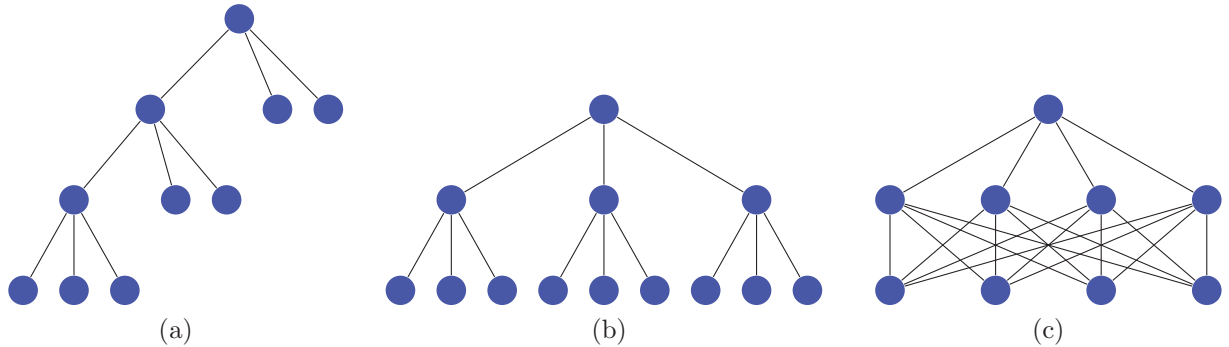

Figure 2: three structures

Figure 3: Illustration of the different multi-level structures. (a) The structure used in the MRA approach. (b) The structure used in the scattering transform approach. (c) The structure used in the convolutional net approach.

In practice, we can also apply some pruning procedure if there are many layers. For example, we can stop expanding the node if it has very small energy.

### 5.3 The convolutional net approach

The structure shown in Figure 2(c) resembles the first few layers of a convolutional net. The root node still represents the input signal, the first layer nodes represent the one-level decomposition coefficients. The coefficients together are then regarded as a multi-channel signal. For example, if the input signal is a monochrome two dimensional image, the first layer coefficients can be regards as a three dimensional image by stacking the $m$ features maps. Once viewed as a three dimensional image, we can construct adaptive frames and bi-frames using three dimensional filters, except that the the filters might not be convolutional in the third dimension since the input signal is not expected to be translation invariant in that direction.

Obviously we are not limited by these three examples of multi-level structures. We call this way of representing the signal *multi-scale adaptive frames and bi-frames*. For convenience we abbreviate it as: *AdaFrame*.

## 6. Examples

### 6.1 The staircase signal

We consider a simple example where the signals are binary, each consists of long sequences of $+1$'s separated by long sequences of $-1$'s, as shown in Figure 4. Let $s$ be the minimum length of consecutive $+1$ and $-1$ blocks. $s$ is a measure of the lowest frequency of the signal. We use Algorithm 1 to learn the filters with $\eta = 10^2$. The filters learned are shown in Figure 5.

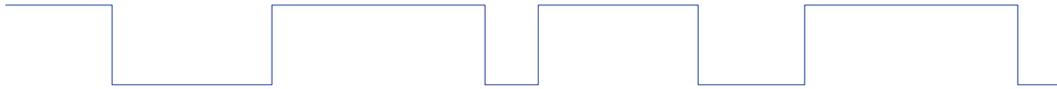

Figure 4: A binary signal

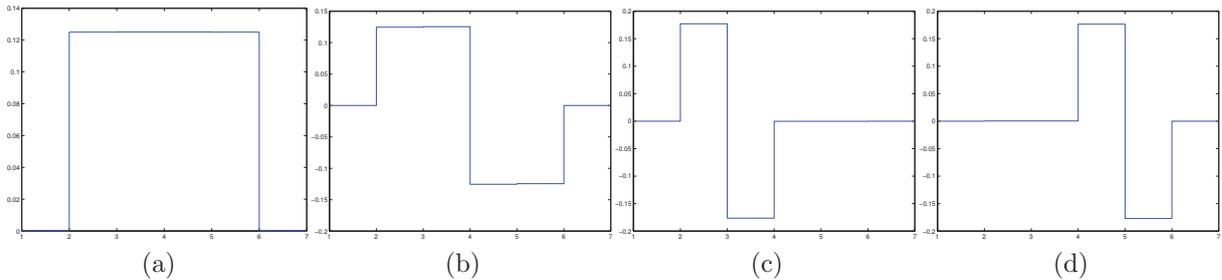

Figure 5: The filters learned using the parameters $m = 4, r = 4, s = 30$.

In the case when $m = 2, r = 2$, we recover the Haar wavelet basis as shown Figure 6.

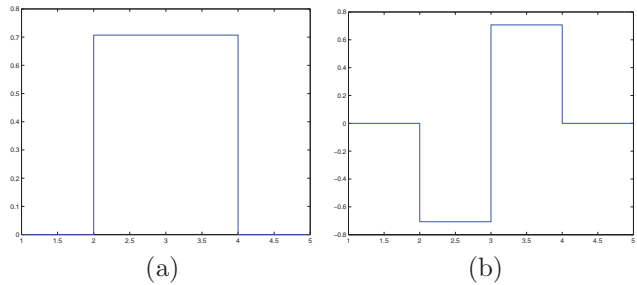

Figure 6: The filters learned using the parameters $m = 2, r = 2, s = 30$. In this case, we recover the Haar wavelets.

This example is simple enough to allow for analytic calculations. In fact, one can show that in the large $s$ regime with the assumption that $r = m$, the filters learned should exactly be the ones shown in the figure. This simple example shows that adaptive filters do capture the special features of the data.

### 6.2 Fingerprint signal

Our next example is the fingerprint dataset Maltoni et al. (2009). We use a fraction of the database. The input are 80 images of size $364 \times 256$. Some sample images and filters learned are shown in Figure 7. The filters are learned using Algorithm 2 with parameters $\eta = 10^2, \lambda = 10^3$. The main feature of the fingerprint images is that they contain oscillations along different directions. As can be seen from the Figure 7, this feature is indeed captured by the learned filters.

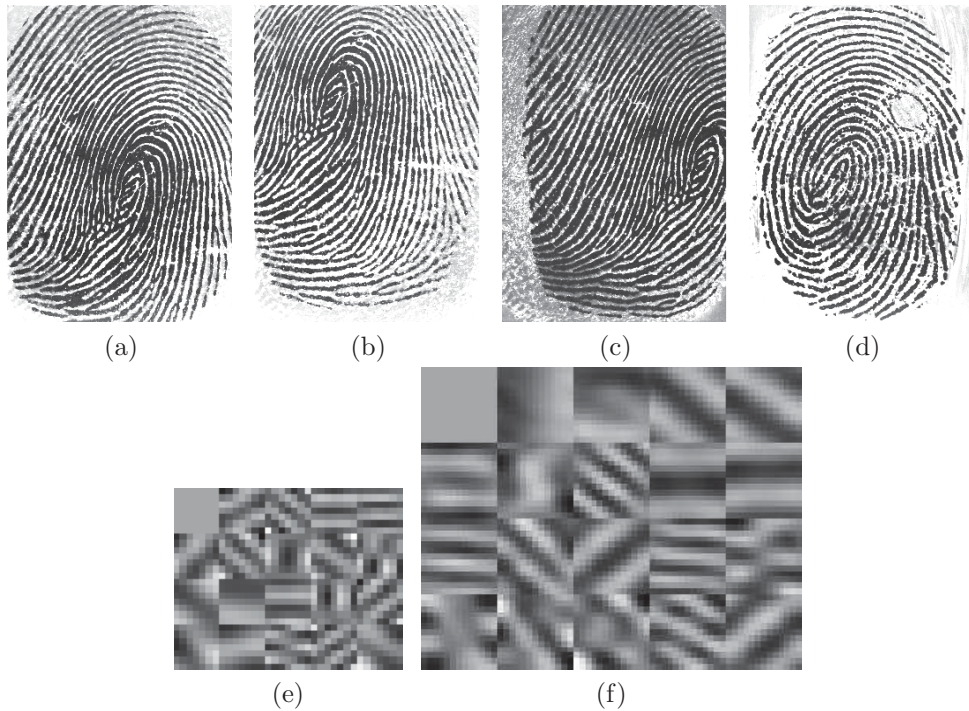

(a)  (b)  (c)  (d)

(e)  (f)

Figure 7: (a)(b)(c)(d)Sample images of finger print.(e) Decomposition filters learned with support size $7 \times 7$. (f) Decomposition filters learned with support size $13 \times 13$.

### 6.3 Another test image

The next example is a well-known natural image shown in Figure 8. This is an example of the redundant bi-frame case. We learn the decomposition filters using Algorithm 2 with $\eta = 10^2, \lambda = 10^3$. Note that some filters look like edge detectors along different directions (e.g. the second and third filters in the first row act like edge detectors along the $x$ and $y$ axis). Most filters look like Gabor wavelets. They detect oscillations along different directions. Because this is an example of the redundant case, hence the reconstruction filters are not unique.

## 7. Recover Predefined Wavelets

The proposed framework is an adaptive extension of the well-known wavelets and wavelet frames. It is natural to ask whether the standard wavelet filters can be recovered using this framework. Naturally we expect that if the signal has a sparse representation in a predefined wavelet domain, then the adaptive frames and bi-frames would recover the predefined wavelets.

To see whether this is the case, we generate the signals using linear combinations of different wavelets with different levels of sparsity. Specifically, the signals are generated using 4 Daubechies wavelets of different support size, "db2","db3","db12","db24" in MATLAB syntax. Sparse random vectors with a given sparsity level are generated (the sparsity level is the ratio of the number of nonzeros coefficients to the length of the coefficient vector, we also call it the density), and these vectors are used as the coefficients of the signals under the wavelet transform.

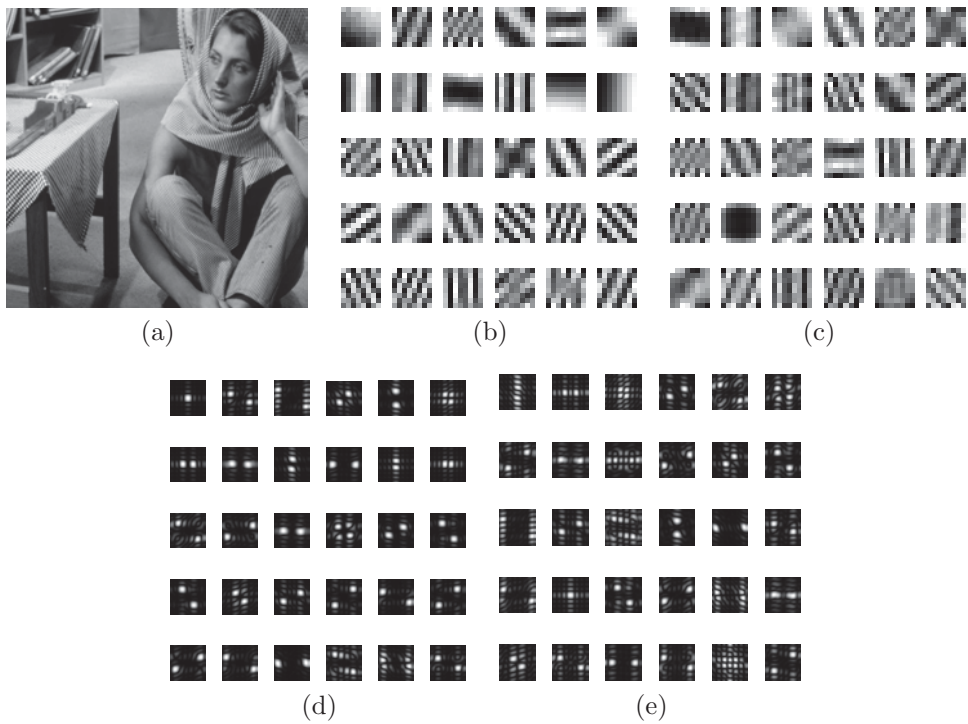

Figure 8: (a) Input image of size $512 \times 512$. (b) 30 decomposition filters with support size $8 \times 8$. (c) A specific set of reconstruction filters. (d) Fourier spectrum of the decomposition filters. (e) Fourier spectrum of the reconstruction filters.

| Density | db2 | db3 | db12 | db24 |
|---|---|---|---|---|
| 0.1 | 1 | 1 | 1 | 1 |
| 0.2 | 1 | 1 | 1 | 1 |
| 0.3 | 1 | 1 | 1 | 1 |
| 0.4 | 1 | 1 | 0 | 0 |
| 0.5 | 0 | 0 | 0 | 0 |

Table 2: Ratio of successful recovery of predefined wavelets.

Given a signal, the adaptive filters are learned by solving (17). Since (17) is nonconvex, to avoid complications coming from local minimum, we used the simulated annealing algorithm to perform the global optimization. We then compare the filters obtained with the original wavelets used to construct the signal. We declare success if the $l_2$ norm of the difference between the adaptive filters and the predefined wavelets is smaller than $10^{-4}$. Table 2 shows the success rate. 10 trials were performed for each case. The result is indeed consistent with our expectation. It is interesting to see that the transition is very sharp.

Figure 9 shows the adaptive filters for the case when the signals are generated using a dense combination of the predefined wavelets. In this case, the predefined wavelets are not optimal, and the signals have a sparser representation under the adaptive filters, as can be seen from Figure 9(c). The $L_1$ norm of the wavelet coefficients is used as a robust measure of sparsity.

## 8. Sample Applications

In this section, we discuss some examples of applications of the multi-scale adaptive frames, the AdaFrames. A thorough comparison of the proposed model and other existing models will be postponed to future publications.

### 8.1 Image Compression

AdaFrames are designed with the objective of making the decomposition coefficients sparse. Therefore they should be naturally suited for image compression tasks. As an intial step, we will compare the performance of AdaFrames with predefined Daubechies wavelets and Haar wavelets. We use the following simple compression scheme: Given an image $x$ and the filters, we perform a decomposition to the coarsest level to get the coefficients, but we keep only the coefficients with relatively large absolute values and set all the other coefficients to 0. The ratio of the total number of coefficients to the number of coefficients kept is called the "compression ratio" (the entropy coding stage is not considered here). We then perform a reconstruction step to get the reconstructed image $\hat{x}$. The quality of the compression is measured by the peak signal-to-noise ratio (PSNR). For monochrome 8 bit image, PSNR is defined as

$$\text{PSNR}(x, \hat{x}) = 10 \log_{10} \frac{255^2}{\frac{1}{MN} \sum_{i=1}^{M} \sum_{j=1}^{N} (\hat{x}(i,j) - x(i,j)^2)} \tag{41}$$

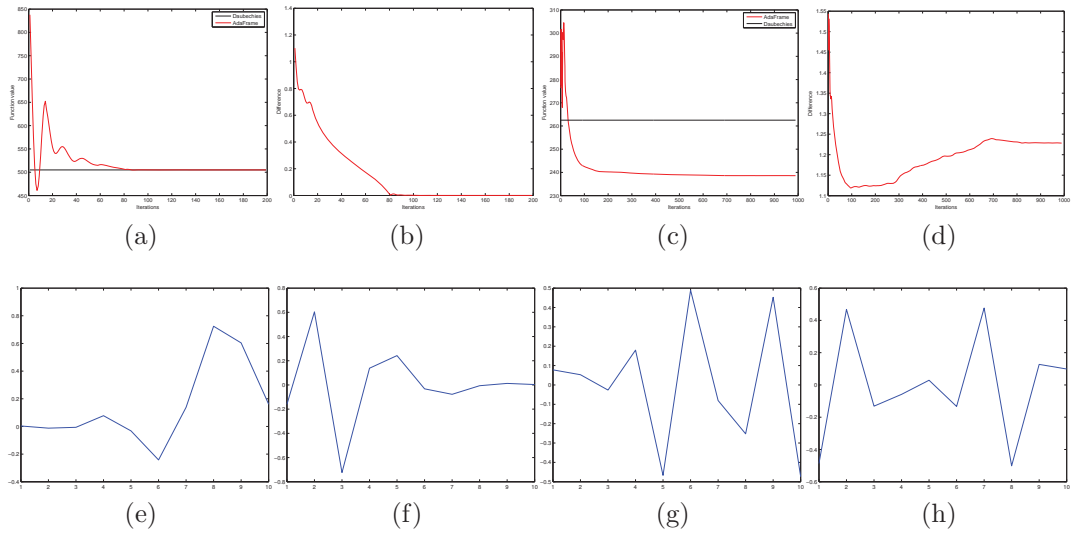

Figure 9: (a) The signal is generated using sparse linear combinations of the Daubechies wavelets. The black line is the objective function value evaluated using the Daubechies wavelets, which is optimal in this case. The value below the black line is due to infeasible intermediate solutions. (b) The filters learned also converge to the Daubechies wavelets, the figure shows the difference of the adaptive filters and the Daubechies wavelets measured in Frobenius norm. (c) The signal is generated using a dense linear combinations of the Daubechies wavelets. In this case, the objective function converges to a value lower than that of the the wavelets, indicated by the horizontal line. (d) The filters learned also converge, but to something different from the Daubechies wavelets. (e)(f) Decomposition filters of the Daubechies wavelet "db5". The signal is generated using sparse linear combination of this wavelets, the filters learned are the same as the wavelets. (g)(h) The filters learned for signals generated using dense combinations of the "db5" wavelet. They are different from the wavelet filters.

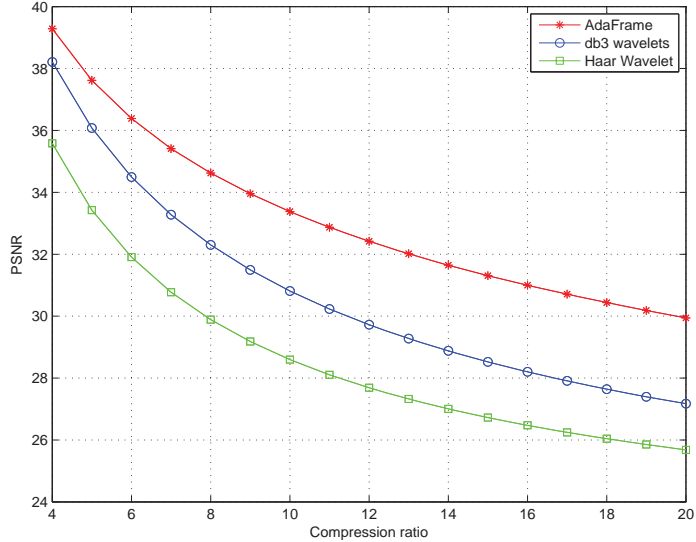

Figure 10: Image compression example. With the same image quality (measured in PSNR), AdaFrames achieves significantly higher compression ratio than the Haar wavelets and the Daubechies wavelets.

The filters are learned using image 8(a). 4 filters of support size $6 \times 6$ are learned using Algorithm 1 with $\eta = 10^2$. The coefficients are critically down-sampled with sampling matrix $M = \text{Diag}(2, 2)$. Initialization is done using the Daubechies filters db3. In general, we have found that using predefined wavelet frames as initialization works quite well. 7 levels of decompositions are performed using the architecture shown in Figure 2 and the same set of filters. The PSNR values are plotted against the "compression ratio" in Figure 10.

## 8.2 Image Denoising

As one of the simplest inverse problem, image denoising provides a convenient platform over which image processing ideas and techniques can be tested. Indeed, during the past few decades, many ideas from a diverse range of viewpoints have been proposed to address this problem, including wavelet domain thresholding, nonlocal means Buades et al. (2005), BM3D Dabov et al. (2007), and the more recent ones based on dictionary learning Elad and Aharon (2006).

Among the various models, we select the K-SVD model Elad and Aharon (2006) as a benchmark for comparison since it is closely related to AdaFrames and since it has been shown to achieve the state of the art results.

Assume the image is corrupted by some additive noise:

$$g = f + n$$

where $f$ is the clean image, $g$ is our observation, and $n$ is the noise with unknown distribution. First let us recall the procedure for wavelet domain denoising. Let $W_A$ and $R_A$ be the decomposition and reconstruction operators associated with the filters $A$ respectively. Given an observed image $x$, the denoised image is then given by:

$$\hat{x} = R_A(\text{shrink}(W_A x)) \tag{42}$$

The procedure for AdaFrame denoising is exactly the same as that of wavelet domain denoising. Given the input image, we first learn the filters from the data using Algorithm 1 (or Algorithm 2 if we want to use bi-frames). We then use (42) to denoise.

In the first example, the input is a single image normalized to $[0, 1]$ and is corrupted with an additive Gaussian white noise with $\sigma = 0.1$. We train the filters both from the noisy image and the clean image with $m = 36, r = 6, \eta = 10^2, \lambda = 10^3$. A two-level decomposition is performed. The soft thresholding parameter is set to be 0.14. Initialization is done by setting the filters to be random orthogonal vectors. The result is shown in Figure 11. The performance of the K-SVD algorithm depends on the number of the atoms in the dictionary. Generally, the performance is better as we increase the number of atoms. In this example, 256 atoms with size $6 \times 6$ are used.

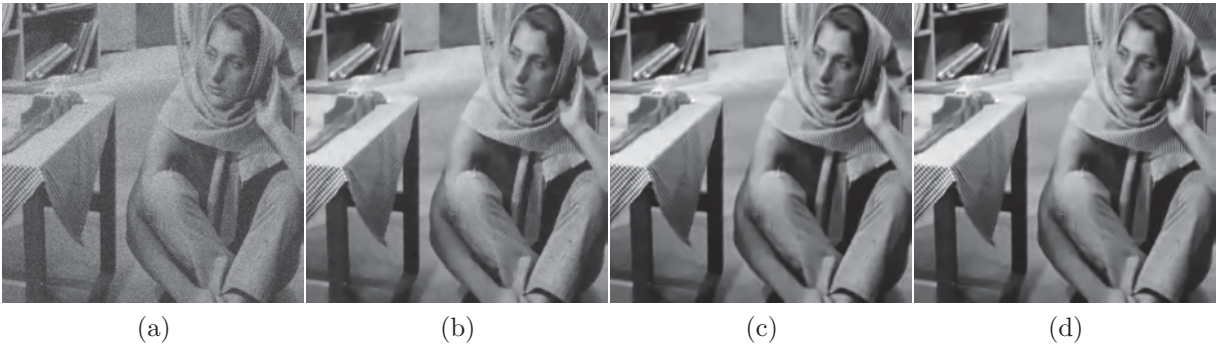

(a) (b) (c) (d)

Figure 11: (a) Noisy input image, $\sigma = 0.1$. (b) K-SVD denoising result, PSNR=28.65dB. (c) AdaFrame denoising, filters learned from noisy image, PSNR=28.8dB. (d) AdaFrame denoising, filters learned from the clean image, PSNR=29.3dB.

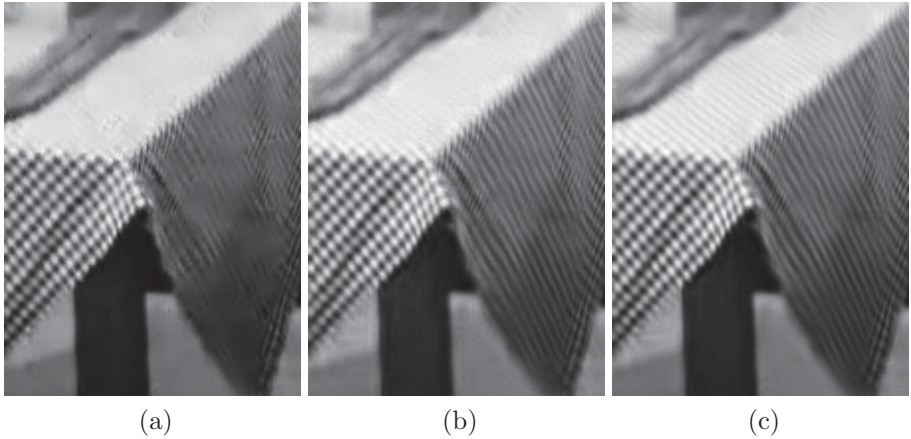

|  (a)  |  (b)  |  (c)  |

Figure 12: (a) Zoom in Figure 11(b). (b) Zoom in of Figure 11(c). (d) Zoom in of Figure 11(d).

It is not surprising that the filters learned from a clean image produces better quality images: One can see from Figure 12 that the fine textures of the image are recovered. At a first sight, one might feel that this is impractical since we normally do not have access to the clean images. Nevertheless, there do exist realistic settings where learning from clean images makes sense. One such a situation is that filters learned from one set of clean images can then be used on another set of noisy images. We tested this idea on the extended Yale human face dataset B Lee et al. (2005). It contains 16128 images of 28 human subjects. We used a subset of the images by picking the first 20 images of each of the subjects. We then added Gaussian white noise with $\sigma = 0.1$ to get the simulated noisy images. A glimpse of the dataset is in Figure 13.

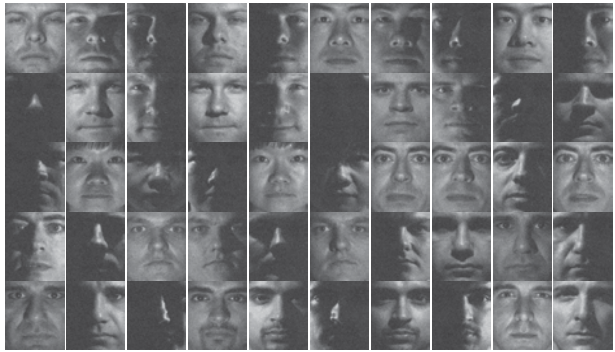

Figure 13: Simulated noisy images from extended Yale face dataset B

To learn the filters, we pick the 100 clean images at random and use Algorithm 2 with $m = 36, r = 6, \eta = 10^2, \lambda = 10^3$. Two-levels of decompositions are performed.

| Test Image | K-SVD $8 \times 8$ | K-SVD $12 \times 12$ | AdaFrame $8 \times 8$ | AdaFrame $12 \times 12$ |
|---|---|---|---|---|
| Barbara $\sigma = 0.02$ | 38.02 | 38.00 | 37.34 | 38.21 |
| Barbara $\sigma = 0.05$ | 33.28 | 33.01 | 31.87 | 33.22 |
| Barbara $\sigma = 0.1$ | 29.47 | 29.24 | 29.18 | 29.70 |
| Boat $\sigma = 0.02$ | 37.02 | 36.71 | 36.75 | 36.86 |
| Boat $\sigma = 0.05$ | 32.53 | 32.11 | 32.50 | 32.59 |
| Boat $\sigma = 0.1$ | 29.19 | 28.70 | 29.18 | 29.21 |
| House $\sigma = 0.02$ | 39.45 | 39.25 | 39.18 | 39.17 |
| House $\sigma = 0.05$ | 35.12 | 34.74 | 34.50 | 34.66 |
| House $\sigma = 0.1$ | 32.15 | 32.05 | 31.19 | 31.45 |
| Lena $\sigma = 0.02$ | 38.45 | 38.21 | 37.98 | 38.45 |
| Lena $\sigma = 0.05$ | 34.46 | 34.18 | 33.21 | 34.34 |
| Lena $\sigma = 0.1$ | 31.38 | 30.84 | 31.12 | 31.39 |
| Peppers $\sigma = 0.02$ | 37.68 | 37.47 | 37.30 | 37.46 |
| Peppers $\sigma = 0.05$ | 33.94 | 33.52 | 33.32 | 33.79 |
| Peppers $\sigma = 0.1$ | 31.26 | 30.78 | 30.33 | 30.91 |

Table 4: Comparison of AdaFrame and K-SVD, performance measured in PSNR, the unit is dB.

The soft-thresholding parameter is set to be 0.14. The results for the noisy images are reported in Table 3.

| | K-SVD, noisy | K-SVD, clean | AdaFrame, noisy | AdaFrame, clean |
|---|---|---|---|---|
| PSNR | 31.4dB | 32.01dB | 31.35dB | 32.07dB |

Table 3: Average PSNR on the simulated noisy images on the extended Yale human face dataset B.

In another experiment, we test the performance of AdaFrame and K-SVD with different support sizes. We use some well-known benchmark images as test images. The images are normalized to $[0, 1]$ and the noise is Gaussian with $\sigma = 0.02, 0.05$ and $0.1$ respectively. For K-SVD, 256 filters of support size $8 \times 8$ and $12 \times 12$ are used. For AdaFrame, 64 filters of support size $8 \times 8$ and 144 filters of support size $12 \times 12$ are learned. $\lambda$ is chosen based on the noise level and is set to be $\lambda = 0.005, 0.01, 0.025$ respectively. The result is shown in Table 5.

As a last denoising example, we apply AdaFrames to some examples of natural photos with unknown noise. The setting is the same as the previous example. We learn filters directly from the noisy images. Since the image has RGB channels, we learn the filters (of support size $9 \times 9$) for each channel seperately with the same value of $\lambda$, which is chosen to yield a good visual impression. The results are shown in Figure 14.

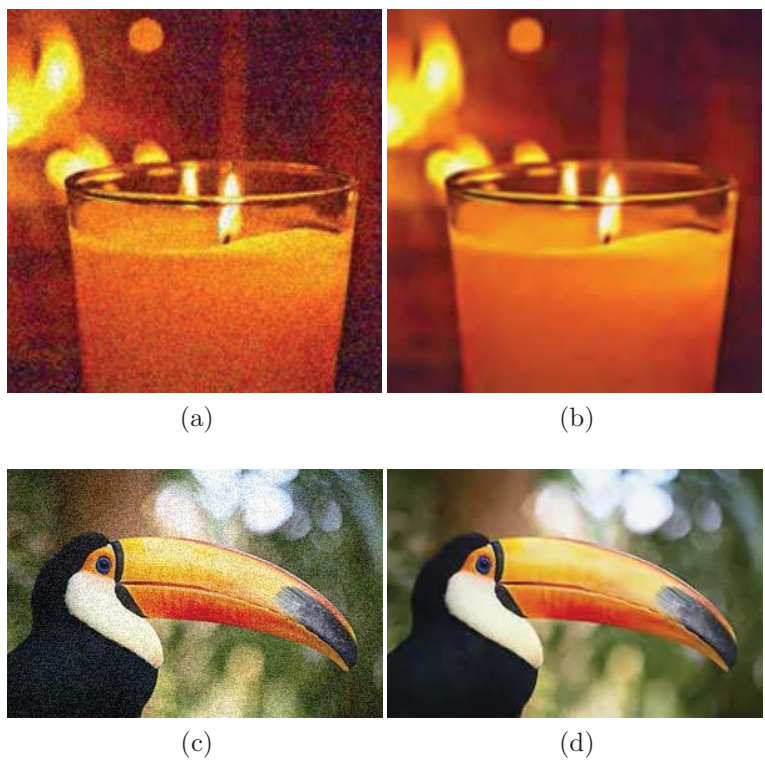

Figure 14: (a)(c) Two images from the Internet. (b)(d) Denoised images using AdaFrame.

As we emphasized earlier, the AdaFrame is faster than sparse coding technique at inference time. We record the computation time for the K-SVD denoising algorithm and the AdaFrame denoising algorithm. In our laptop with the same setup, the K-SVD algorithm takes 25s to train a dictionary with 256 atoms of support size $8 \times 8$ and 6.5s to denoise the image. The software we use is downloaded from http://www.cs.technion.ac.il/~ronrubin/software. The AdaFrame takes 3.7s to train 64 filters with support size $8 \times 8$ and takes 0.6s to denoise. The time for denoising scales linearly with the number of filters.

### 8.3 Image Classification

Although AdaFrames are aimed to produce sparse representations, they can also be used to for other tasks such as extracting features for object recognition. In fact, it can provide a faster alternative to sparse coding.

To demonstrate this idea, in the following example, we apply AdaFrames to extract features in order to classify the handwritten digits. The dataset we used is MNIST LeCun et al. (1998). It contains 70000 $28 \times 28$ images of digits from 0 to 9, 60000 for training and 10000 for testing. A nonlinear transformation, the rectified linear function defined by $relu(x) = max(x, 0)$ is applied to the coefficients obtained using

AdaFrames. The results are sent to a linear support vector machine (SVM) to perform the classification task. We discuss three different set of experiments.

In the first setup, we use Algorithm 2 to learn the filters with $m = 6, r = 6, \eta = 10^2, \lambda = 10^3$. Initialization is done with random orthogonal filters. For each image, we perform a one-level decomposition to get the coefficients.

The second setup is identical to the first one, except $m = 12$ instead of $m = 6$. It is generally believed that lifting the raw pixels to some higher-dimensional feature space will be helpful for classification. Since we use more filters in this setup, the features we get have higher dimensions. Indeed the results are better than the results of the previous setup.

In the third setup, we use a two-level decomposition. We use Algorithm 2 to learn the filters with $m = 6, r = 6, \eta = 10^2, \lambda = 10^3$. Same nonlinear transformation as in the previous setups are used. In this way, we obtain 6 feature maps, each of size $28 \times 28$. Then the collection of the feature maps are treated as 6 sets of new input images. For each set, we use Algorithm 2 with $m = 4, r = 6, \eta = 10^2, \lambda = 10^3$ to learn the filters. Hence we have 24 filters in total. For each feature map, we perform a one-level decomposition using the corresponding 4 filters to get 4 feature maps. Again, we keep the positive coefficients and set the negative coefficients to 0. These positive coefficients in the first and second layers are the extracted features.

| MNIST | Raw pixel | I | II | III |
|---|---|---|---|---|
| Precision | 88.0 % | 97.0 % | 97.4% | 99.0% |

Table 5: Results of the MNIST classification. "Raw pixel" means that the features are the raw pixels.

These features are sent to a linear SVM. The results are reported in Table 5. Note that there is a significant reduction in the error rates compared to raw pixel features. As a point of comparison, the state-of-the-art result with preprocessing, is 0.23%, which is obtained using deep convolutional neural networksCiresan et al. (2012).

## 9. Connection with De-convolutional net

Convolutional nets have had remarkable successes in a variety of challenging applications LeCun et al. (1998); Lee et al. (2009); Krizhevsky et al. (2012). A typical supervised convolutional net consists of several convolutional layers and fully connected layers. A convolutional layer has the structure shown in Figure 15. It maps the feature maps produced by the previous layer to another set of feature maps. The input feature maps are first convolved with some filters, which are also obtained from training. A point-wise nonlinear function, called the "activation function", such as a rectified linear function is then applied, followed by a pooling procedure in order to down-sample the set of feature maps. Pooling is usually a local operation. Max pooling, namely picking the feature map with the maximum amplitude in a small neighborhood of each node, is the most popular. It is similar to simple down-sampling but is nonlinear.

Although convolutional nets are designed for feature extraction and object recognition, it is an interesting question to ask how much of the input data can be reconstructed from the information in the intermediate layers of the network. For one thing, this can help us to gain some intuition about how convolutional nets work.

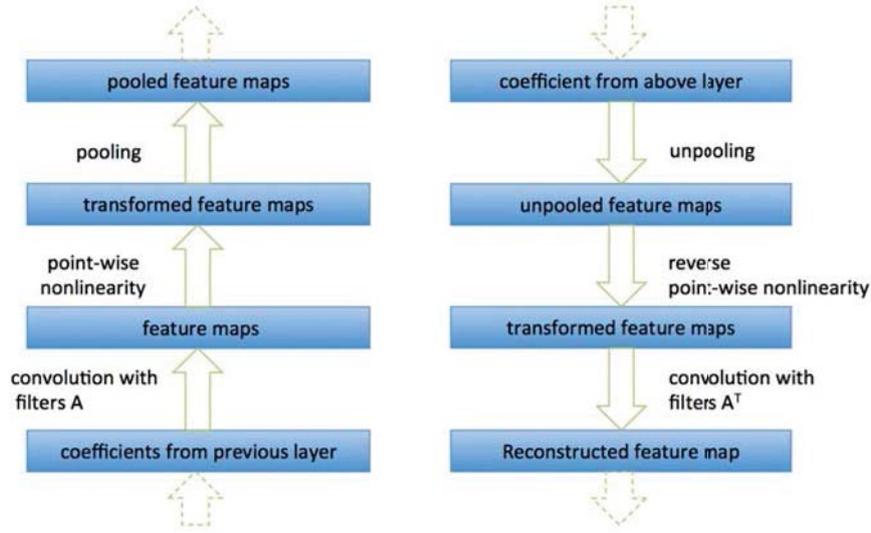

Figure 15: The left figure shows the typical structure of a convolutional layer from a convolutional net, the right figure shows the structure of a de-convolutional layer from a de-convolutional net.

In this regard the most popular approach in the literature is the "deconvolutional net" Zeiler et al. (2010). A deconvolutional net can be thought of as a convolutional net that uses the same components (filtering, nonlinear activation, pooling) but in reverse order. Specifically a deconvolutional net consists of the following steps: First, the pooling procedure is reversed. If averaging or other linear operator is used for pooling, then to reverse it, one simply applies its transpose operator. The max-pooling procedure is a non-linear operation. For an image $I$, the max-pooling operation has two outputs, the maximum value and the position where the maximum value is obtained, defined as

$$(v, p)(x) = (\text{sign}(I(x)) \cdot \max_{x \in \mathcal{N}} |I(x)|, \arg\max_{x \in \mathcal{N}} |I(x)|)$$

where $\mathcal{N}$ is the neighborhood of $x$. To reverse max-pooling, we set

$$I(x) = \begin{cases} v & : x = p, x \in \mathcal{N} \\ 0 & : x \neq p, x \in \mathcal{N} \end{cases}$$

The second component is to reverse the activation function. For invertible functions such as the sigmoid or the tanh function LeCun et al. (1998), we simply take their inverse. The situation where the activation function is non-invertible as is the case of the absolute value function is more complicated and is discussed in Waldspurger et al. (2012).

The third component is to reverse the convolution operator, hence the name "deconvolution net". Since convolution is a linear operator, to reverse it, one applies its transpose Zeiler and Fergus (2013).

The above procedure is summarized in a diagram in Figure 15. Notice the similarity with applying wavelet frame transforms. A single level decomposition and reconstruction step of the wavelet frame transform can be described as in Figure 16. We see that if we ignore the point-wise nonlinearity, a convolutional or a deconvolutional layer is very similar to a decomposition and reconstruction step in wavelet bi-frame transform respectively.

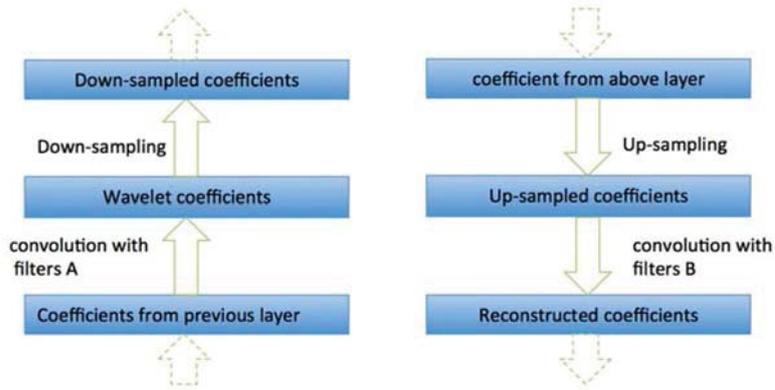

Figure 16: One level decomposition and reconstruction of AdaFrame

There is a subtle but important difference. In deconvolutional net, deconvolution is done by applying the transpose of the convolution operator. In the one level wavelet bi-frame reconstruction, this is done using the reconstruction filters, obtained by solving (10), as required by UEP. Since there is no guarantee that the UEP condition is satisfied by the filters obtained in the convolutional nets, one expects that there will be errors in the reconstruction process, i.e. the deconvolutional nets. This is indeed the case, as we show below.

The similarity between the convolutional layer and one level wavelet frame transform suggests a natural fix for this problem. Instead of using the flipped convolutional filters as the deconvolutional filters, we view the convolutional filters as the decomposition filters and solve (10) to obtain the reconstruction filters. These reconstruction filters are then used as the deconvolutional filters. Everything else is the same as in the original deconvolutional net. The existence of a solution to (10) is guaranteed by the fact that in a typical convolutional net, the number of filters is large, and hence we are in the the redundant case for the wavelet bi-frames. This small change to the deconvolutional net yields much better reconstruction as we now demonstrate.

We implemented a two-layer convolutional network. In the first layer, we have 12 filters of support size $6 \times 6$, the pooling procedure is chosen to be the usual down-sampling with decimation factor $(2, 2)$. To construct the second layer, we stack together the feature maps from the first layer and form a three-dimensional signal. We then learn 12 filters of support size $4 \times 4 \times 2$, the pooling procedure is also down-sampling with decimation factor $(2, 1, 2)$. The activation function is the sigmoid function. The results of reconstructing the input image using the original deconvolutional net and the modified procedure described above are shown in Figure 17. As one can see, using the deconvolutional net approach, we gradually lose information as we ascend in the layers, while using the AdaFrame, we do not lose information.

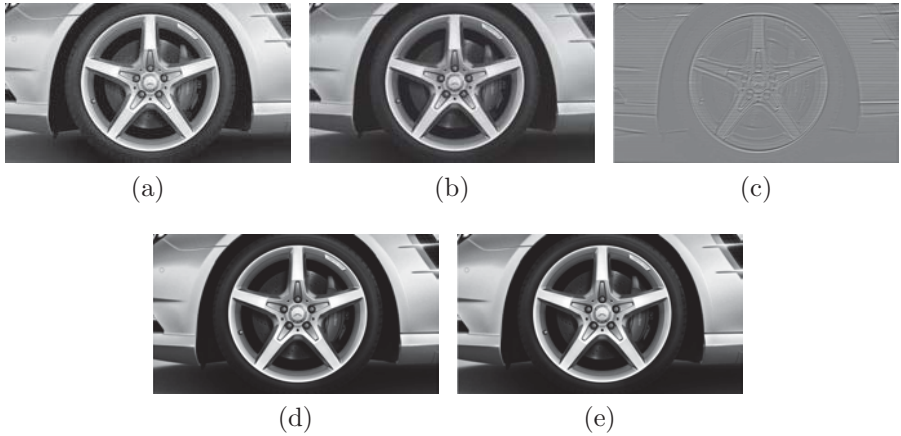

(a) (b) (c)

(d) (e)

Figure 17: (a) The input image. (b) Reconstruction from the first layer activations using "deconvolutional net" approach. (c) Reconstruction from the second layer activations using the "deconvolutional net" approach. (d) Reconstruction from the first layer activations using the AdaFrame. (e) Reconstruction from the second layer activations using the AdaFrame.

In addition to near perfect reconstruction, AdaFrame has the potential to be used as an initialization method for the convolutional parts of a typical convolutional net. This is a direction for future research.

## 10. Conclusion

Predefined wavelets and dictionary learning have both been very successful in their own ways. In this paper, we have proposed a framework, the AdaFrame, that naturally combines the advantages of both. It is multi-scale and computationally efficient as pre-defined wavelets and wavelet frames, while being adaptive as in dictionary learning. Unlike dictionary learning, the proposed framework guarantees perfect reconstruction, which is an appealing property in many signal processing tasks.

Between adaptive frames and adaptive bi-frames, our experience suggests that adaptive bi-frames are much easier to use because of the additional flexibility. The learning procedure is also easier, especially when the system is very redundant in which case the learning procedure can be carried out in two phases by learning the decomposition and reconstruction filters separately.

In addition to the examples given in this paper, we believe that the proposed framework can be useful in many other applications. It is not restricted to image processing, it can be used on time series, videos and even graphs. We will explore these applications in subsequent papers.

Another direction for future investigation is to use the proposed framework as feature extraction tools for machine learning tasks. Sparse coding has been popular for this purpose. But the proposed framework should be a promising alternative since it is more efficient and it has a multi-scale structure. It should be particularly appealing when the computation cost is the main bottleneck, as is the case in some real-time object recognition systems.

## Acknowledgments

This work is supported in part by the 973 program of the Ministry of Science and Technology of China, the Major Program of NNSFC under grant 91130005 and an ONR grant N00014-13-1-0338.

# Appendix

**Proof of Theorem 1**

For convenience, we need the following lemma.

**Lemma 2** *Let $M$ be $d \times d$ sampling matrix and $a, b \in l_2(\mathbb{Z}^d)$ be finitely supported sequences. Then*

$$\widehat{\mathcal{S}_b v}(\xi) = |\det(M)| \hat{v}(M^T \xi) \hat{b}(\xi) \tag{43}$$

*and*

$$\widehat{\mathcal{T}_a}(M^T \xi) = |\det(M)|^{-1} \sum_{\omega \in \Omega_M} \hat{v}(\xi + 2\pi\omega) \overline{\hat{a}(\xi + 2\pi\omega)} \tag{44}$$

*where*

$$\hat{a}(\xi) := \sum_{k \in \mathbb{Z}^d} a(k) e^{-ik \cdot \xi}$$

*and*

$$\Omega_M := [(M^T)^{-1} \mathbb{Z}^d] \cap [0, 1)^d$$

**Proof** For a sequence $v \in l_2(\mathbb{Z}^d)$,

$$\begin{aligned}
\widehat{\mathcal{S}_b v}(\xi) &= \sum_{k \in \mathbb{Z}^d} (S_b v)(k) e^{-ik \cdot \xi} \\
&= |\det(M)| \sum_{k \in \mathbb{Z}^d} \sum_{j \in \mathbb{Z}^d} v(j) b(k - Mj) e^{-ik \cdot \xi} \\
&= |\det(M)| \sum_{k \in \mathbb{Z}^d} b(k - Mj) e^{-i(k - Mj) \cdot \xi} \sum_{j} v(j) e^{-iMj \cdot \xi} \\
&= |\det(M)| \hat{b}(\xi) \hat{v}(M^T \xi).
\end{aligned} \tag{45}$$

Let $\hat{u}(\xi) = \sum_{k \in \mathbb{Z}^d} v(k) \overline{a(k-n)}$, then $\hat{u}(\xi) = \hat{v}(\xi) \overline{\hat{a}(\xi)}$. By definition of $\mathcal{T}_a$, we have $(\mathcal{T}_a)(n) = u(Mn)$. So

$$\widehat{\mathcal{T}_a v}(M^T \xi) = \sum_{n \in \mathbb{Z}^d} (\mathcal{T}_a v)(n) e^{-in \cdot M^T \xi} = \sum_{n \in \mathbb{Z}^d} u(Mn) e^{-kMn \cdot \xi} \tag{46}$$

On the other hand,

$$\begin{aligned}
\sum_{\omega \in \Omega_M} \hat{u}(\xi + 2\pi\omega) &= \sum_{k \in \mathbb{Z}^d} \sum_{\omega \in \Omega_M} e^{-ik \cdot (\xi + 2\pi\omega)} \\
&= \sum_{k \in \mathbb{Z}^d} u(k) e^{-ik \cdot \xi} \sum_{\omega \in \Omega_M} e^{-ik \cdot 2\pi\omega}.
\end{aligned} \tag{47}$$

If $k \in M\mathbb{Z}^d$, then $\sum_{\omega \in \Omega_M} e^{-ik \cdot 2\pi\omega} = |\det(M)|$; if $k \in \mathbb{Z}^d / M\mathbb{Z}^d$, $\sum_{\omega \in \Omega_M} e^{-ik \cdot 2\pi\omega} = 0$, so we have

$$\sum_{\omega \in \Omega_M} \hat{u}(\xi + 2\pi\omega) = |\det(M)| \sum_{k \in M\mathbb{Z}^d} u(k) e^{-ik \cdot \xi} = |\det(M)| \sum_{n \in \mathbb{Z}^d} u(Mn) e^{-iMn \cdot \xi}. \tag{48}$$

Combining this with (46), we get the desired result. ∎

**Lemma 3** *Let $M$ be $d \times d$ sampling matrix and $a_l, b_l, l = 1, \cdots, m$ be $m$ finitely supported sequences. Then*

$$\sum_{l=1}^{m} \mathcal{S}_{b_l} \mathcal{T}_{a_l} v = v, \quad \forall v \in l_2(\mathbb{Z}^d) \tag{49}$$

*if and only if, for any $\omega \in \Omega_M := [(M^T)^{-1}\mathbb{Z}^d] \cap [0,1)^d$*

$$\sum_{l=1}^{m} \hat{b}_l(\xi)\overline{\hat{a}_l(\xi + 2\pi\omega)} = \delta(\omega). \tag{50}$$

**Proof** By definition of the decomposition and reconstruction operators $W_a$ and $R_b$, we have

$$R_b W_a v = \sum_{l=1}^{m} \mathcal{S}_{b_l} \mathcal{T}_{a_l} v. \tag{51}$$

which is equivalent to

$$\sum_{l=1}^{m} \widehat{\mathcal{S}_{b_l} \mathcal{T}_{a_l} v}(\xi) = \hat{v}(\xi), \forall v \tag{52}$$

By the above lemma, we have

$$\begin{aligned}
\hat{v}(\xi) &= \sum_{l=1}^{m} (\widehat{\mathcal{S}_{b_l} \mathcal{T}_{a_l} v})(\xi) \\
&= \sum_{l=1}^{m} |\det(M)| \widehat{\mathcal{T}_{a_l} v}(\xi)(M^T \xi)\hat{b}_l(\xi) \\
&= \sum_{l=1}^{m} \sum_{\omega \in \Omega_M} \hat{v}(\xi + 2\pi\omega)\hat{b}_l(\xi)\overline{\hat{a}_l(\xi + 2\pi\omega)}
\end{aligned} \tag{53}$$

If (52) holds true, then

$$\sum_{l=1}^{m} \sum_{\omega \in \Omega} \hat{v}(\xi + 2\pi\omega)\hat{b}_l(\xi)\overline{\hat{a}_l(\xi + 2\pi\omega)} = \sum_{\omega \in \Omega_M} \hat{v}(\xi + 2\pi\omega)\delta(\omega) = \hat{v}(\xi). \tag{54}$$

holds for all $v \in l_2(\mathbb{Z}^d)$.

Conversely, if (51) is true, we can choose $v$ that is close to a $\delta$-function. Let $B_\epsilon(\xi_0)$ be the open ball centered at $\xi_0$ with radius $\epsilon$. Fix $\omega_0 \in \Omega_M$ and $\xi_0 \in \mathbb{R}^d$, we can choose $v \in l_2(\mathbb{Z}^d)$ such that

1. $\hat{v}(\xi + 2\pi\omega_0) = 1$, for all $\xi \in B_\epsilon(\xi_0)$.
2. $\hat{v}(\xi + 2\pi\omega) = 0$, for all $\xi \in B_\epsilon(\xi_0), \omega \in \Omega/\{\omega_0\}$.
3. $supp(\hat{v}) \subset 2\pi\omega_0 + B_{2\epsilon}(\xi_0)$

This is possible because the set $\Omega$ is discrete. ∎

Hence, for $\xi \in B_\epsilon(\xi_0)$,

$$\hat{v}(\xi) = \sum_{l=1}^{m} \sum_{\omega \in \Omega_M} \hat{v}(\xi + 2\pi\omega)\hat{b}_l(\xi)\overline{\hat{a}_l(\xi + 2\pi\omega)} \\ = \sum_{l=1}^{m} \hat{b}_l(\xi)\overline{\hat{a}_l(\xi + 2\pi\omega_0)} \tag{55}$$

Hence,

$$\sum_{l=1}^{m} \hat{b}_l(\xi)\overline{\hat{a}_l(\xi + 2\pi\omega_0)} = \delta(\omega)$$

for all $\xi \in B_\epsilon(\xi_0)$, since $\xi_0$ and $\omega_0$ are arbitrary, we obtain the desired result.

Proof of Theorem 1.

**Proof** We only need to establish that (10) is equivalent to (52).

$$\delta(\omega) = \sum_{l=1}^{m} \sum_{k \in \mathbb{Z}^d} \overline{b_l(k)} e^{ik\cdot\xi} \sum_{n \in \mathbb{Z}^d} a_l(n) e^{-in\cdot(\xi+2\pi\omega)} \\ = \sum_{l=1}^{m} \sum_{k,n \in \mathbb{Z}^d} \overline{b_l(k)} a_l(n) e^{i(k-n)\cdot\xi} e^{-in\cdot 2\pi\omega} \tag{56}$$

Denote by $\Gamma_M := (M[0,1)^d) \cap \mathbb{Z}^d$, then we have $\mathbb{Z}^d = \Gamma_M + M\mathbb{Z}^d$, replace $n$ by $Mn+\gamma$, we can rewrite the above equation as

$$\delta(\omega) = \sum_{l=1}^{m} \sum_{\gamma \in \Gamma_M} \sum_{k,n \in \mathbb{Z}^d} \overline{b_l(k)} a_l(Mn+\gamma) e^{i(k-Mn-\gamma)\cdot\xi} e^{-i(Mn+\gamma)\cdot 2\pi\omega} \\ = \sum_{l=1}^{m} \sum_{\gamma \in \Gamma_M} \sum_{k,n \in \mathbb{Z}^d} \overline{b_l(k+Mn+\gamma)} a_l(Mn+\gamma) e^{ik\cdot\xi} e^{-k\gamma\cdot 2\pi\omega} \tag{57}$$

Note that $(e^{-i\gamma\cdot 2\pi\omega})_{\omega \in \Omega_M, \gamma \in \Gamma_M}$ is the Fourier matrix, and its inverse matrix is $|\det(M)|^{-1}(e^{i\gamma\cdot 2\pi\omega})_{\omega \in \Omega_M, \gamma \in \Gamma_M}$. Therefore,

$$(\sum_{l=1}^{m} \sum_{k,n \in \mathbb{Z}^d} \overline{b_l(k+Mn+\gamma)} a_l(Mn+\gamma) e^{ik\cdot\xi})_{\gamma \in \Gamma_M} \\ = |\det(M)|^{-1}(e^{i\gamma\cdot 2\pi\omega})_{\omega \in \Omega_M, \gamma \in \Gamma_M} (\delta(\omega))_{\omega \in \Omega} \\ = |\det(M)|^{-1}(1,1,\cdots,1)^T. \tag{58}$$

Hence
$$\sum_{l=1}^{m}\sum_{k,n\in\mathbb{Z}^d}\overline{b_l(k+Mn+\gamma)}a_l(Mn+\gamma)e^{ik\cdot\xi}=|\det(M)|^{-1},\quad\forall\gamma \qquad(59)$$

taking inverse Fourier transform, we get the desired result. ∎